\documentclass[11pt,a4paper]{article}
\usepackage[hyperref]{eacl2021}
\usepackage{times}
\usepackage{latexsym}
\usepackage{subfig}
\usepackage{graphicx}
\usepackage{tabularx}
\usepackage{amsmath}
\usepackage{microtype}
\usepackage{multirow}
\usepackage{makecell}

\usepackage{microtype}

\aclfinalcopy 
\title{ A Case Study of Deep Learning Based Multi-Modal Methods for Predicting the Age-Suitability Rating of Movie Trailers}

  \author{Mahsa Shafaei*, Christos Smailis*, Ioannis A. Kakadiaris, Thamar Solorio \\
  Department of Computer Science, University of Houston, Houston, TX, USA
  
  }

\date{}

\begin{document}
\maketitle
\begin{abstract}
In this work, we explore different approaches to combine modalities for the problem of automated age-suitability rating of movie trailers. %conduct a case study of methods for deep learning systems, while addressing the problem of automated age-suitability rating of movie trailers. 
%Specifically, we explore the use of late fusion, feature concatenation fusion and Gated Multi-modal Unit (GMU) based fusion to combine information from the video, subtitles, and audio modalities. 
First, we introduce a new dataset containing videos of movie trailers in English downloaded from IMDB and YouTube, along with their corresponding age-suitability rating labels. Secondly, we propose a multi-modal deep learning pipeline addressing the movie trailer age suitability rating problem. %Finally, our experimental results demonstrate that multi-modal fusion significantly boosted weighted averaged F1 score performance against the best performing individual modality baseline. 
This is the first attempt to combine video, audio, and speech information for this problem, and our experimental results show that multi-modal approaches significantly outperform the best mono and bimodal models in this task. %GMU and late fusion were found to be more effective ways to combine multi-modal information for this problem, showing statistically significant difference from the results produced by feature concatenation fusion.
\end{abstract}

\section{Introduction}
{\let\thefootnote\relax\footnote{{* These authors contributed equally to this work}}}
Movie trailers can be found in abundance throughout the web using services such as video streaming platforms. However, not all types of content in trailers are suitable for every audience. Specifically, movie trailers may contain explicit, aggressive, or violent content that may be harmful to the psyche of young viewers. Previous research has documented that some of the negative effects of mass media in young viewers include aggression and anxiety \cite{wilson2008media,chang2019effect}, as well as increasing the risk of sexual onset and alcohol and drug consumption, unwanted pregnancies, and sexually transmitted diseases \cite{strasburger1989adolescent}.

The Advertising Administration of the Motion Picture Association of America (“MPAA”) established guidelines for manually rating the age-suitability of movie trailers \cite{Advertising_handbook}. The rating of movie trailers is independent of the rating of the movie itself, as a trailer includes only a short overview of the entire movie.  %The MPAA rating procedure is time-consuming and does not scale to an ever-increasing number of trailers available on online streaming platforms
Due to the time consuming nature, as well as the challenges to scale the MPAA rating process, automating the task is of practical value. Moreover, automating the task poses interesting challenges to multi-modal classification systems, as the source of the objectionable content can come from any, or the combination of, these sources: language (use of bad words or discussion of adult themes), images (graphic violent scenes, nudity, drug or alcohol use), and audio (loud noises and music score denoting suspenseful content). A successful rating approach should integrate evidence provided by the multiple modalities when making the predictions.

In this paper, we study the performance of different multi-modal deep learning methods, to automatically predict the MPAA age-suitability rating of movie trailers using cues from the video, audio, and text modalities. Our goal is to show the feasibility of automating the rating task, and in particular, the relevance of multimodal solutions. We explore the use of late fusion, feature concatenation fusion and Gated Multi-modal Unit (GMU) Fusion \cite{arevalo2017gated}. Since the proposed pipeline does not use any type of metadata for trailers or movies, it can be easily extended to be applied to any type of online video content. The main contributions of this work are:
(i) we introduce a new task in multi-modal classification; rating videos based on the MPAA rating metric for movie trailers; (ii) we introduce the Multi-modal Movie Trailer Rating (MM-Trailer) dataset that contains movie trailers and their corresponding MPAA tags, audio files, subtitles of the trailers, and the metadata of the target movie; and (iii) we demonstrate empirically that combining the different modalities yields significant improvements over the strongest monomodal model. Our results show that both, the GMU and late fusion approaches yield promising results.

\section{Related Work}

This work is related to four different areas, namely: (i) text classification, (ii) video classification, (iii) audio classification,  and (iv) movie classification Datasets.  

\noindent\textbf{Text Classification:}
% In \cite{wehrmann2018self} the authors proposed a deep learning architecture for predicting the genre of movies using the plot synopsis. Their model used an attention mechanism to learn the importance of features from each word present in a synopsis. Similarly in \cite{8334466} the authors predicted the movie genres from plot summaries using a Bidirectional LSTM architecture. 
In \cite{martinez2019violence} the authors proposed an RNN-based architecture for detecting violence in movies on a segment level as well as the full movie level, by using the movie's script. In \newcite{shafaei2019rating} the authors proposed an RNN-based architecture with an attention mechanism that jointly models the genre and the emotions in movie script to predict the MPAA rating of a full movie. The main difference between our work and aforementioned papers is that they only use scripts to predict the movie ratings (violence rating and MPAA ratings), while we employ various modalities (audio, video, and text) to predict if a trailer (not the entire movie) is appropriate or not for children. It should be noted that the rating schema is different for trailers compared to movies (details in Section \ref{Datase}), and movies are not freely available on the internet.

\noindent\textbf{Video Classification:} Early approaches, such as \cite{Karpathy} on video classification using Deep Learning, explored the use of several temporal fusion methods for combining information from multiple consecutive video frames using features extracted from CNN architectures. The authors in \cite{lrcn2014} introduced an end-to-end architecture based on a combination of CNNs used for feature extraction from RGB frames. The CNN features are then forwarded to an LSTM layer that models the temporal variation of frames. A different approach is followed in \cite{3DCNN}, namely 3D-CNN, where authors propose the use of a CNN variant that takes into account convolutions performed into both the spatial and temporal domains of a video. An expansion of the 3D-CNN approach was proposed by \cite{QuoVantis}, where the authors propose a two-stream 3D-CNN architecture for video classification. Again the two streams used as input RGB frame data and Optical flow images. 

\noindent\textbf{Audio Classification:}
In past research, several types of handcrafted feature extraction techniques have been proposed for the audio modality \cite{MFCCs,zero_Crossing,papakostas2017deep} with the ones being the most prominently used in the literature being Mel-frequency cepstral coefficients (MFCCs).
However, recently several approaches have been proposed for combining audio features such as spectrogram information with deep learning architectures to perform audio classification \cite{papakostas2017deep,45611,koutini2019receptive}. Audio has been explored as a modality for classifying movie content in several works such as \cite{1048494, hebbar2018improving}. However, none of these methods has focused on the problem of movie trailer age-suitability rating.

% \noindent\textbf{Multi-modal Data Fusion:}
% Multi-modal data fusion is an attempt to integrate information from different resources to predict an output (class label or real value). Fusion methods can be categorized into three general groups \cite{baltruvsaitis2018multimodal}; early, late, and hybrid. In early fusion, low-level features are combined and fed to the prediction model. However, in late fusion,  different modalities are merged in the decision level using various rules (e.g., majority voting, averaging). Hybrid methods have the advantages of both methods.
% In \cite{10.1007/978-3-319-23989-7_8} the authors introduced a late fusion scheme based on SVM classifiers and handcrafted features to perform movie genre classification using information from plot synopsis and movie posters. In \cite{arevalo2017gated}, the authors introduced a deep learning model based on gated multi-modal units, that can be categorized as a hybrid method, for genre prediction, fusing movie image posters, and movie textual plot information. 

\noindent\textbf{Movie Classification Datasets:} Several movie classification datasets have been proposed in the past. In \cite{demarty2014benchmarking}, the authors introduced MediaEval 2013 Violent Scene Detection, which provided annotations for detecting violent scenes in movies. In \newcite{9064936}, the authors proposed an evaluation framework, for Violent Scenes Detection in Hollywood and YouTube videos along with a dataset (VSD96). Although these datasets are relevant to our work, they only cover the violence aspect and cannot address the problem of age-suitability rating (violence is only one of many aspects of age rating). In \cite{shafaei2019rating} the authors proposed a movie dataset focusing on the task of predicting the MPAA rating of the movie. However, the aforementioned dataset only includes movie scripts and corresponding metadata but does not include movie trailers or related age-suitability tags (As we mentioned earlier, the MPAA rating scheme is different for movies and trailer). In \cite{cascante2019moviescope}, authors introduced Moviescope, a dataset for movie genre classification. Similarly, it does not include MPAA age-suitability rating labels for movie trailers.

\section{Dataset}
\label{Datase}
To the best of our knowledge, there is no previous trailer dataset with age-suitability rating. Thus, we assembled the multi-modal Movie Trailer dataset (MM-Trailer)\footnote{We will release the dataset for the public access when anonymity of authors is not a concern} by collecting the rated trailers from the IMDB website and YouTube. Typically trailers are advertising movies soon to be released and shown in theaters before a movie starts. Rating in trailers is shown by a colored band (red, yellow, green) and a message that appears at the beginning of the trailer. The rating of the trailers adheres to the rating of the movie being shown in the theater. For instance, if the movie playing at the theater is rated as NC-17 (no one under 17 is recommended to watch this movie), the green band trailer that is advertised before this movie may not be appropriate for children even if the color is green. The yellow band is designed for trailers advertised on the internet, and it indicates that the corresponding trailer is suitable for ``age-appropriate internet users'' as visitors to sites are mainly adults. The last group of trailers are red band trailers; red color indicates the content is only appropriate for a ``mature audience'' or ``restricted audience''. 

Since our goal is to design an automated system that is able to predict which movie trailers are not recommended for children, we define only two classes of trailers for the dataset: 
\begin{enumerate}
\item \textbf{Green-band trailers:} this category includes (i)  trailers with the message ``all audiences'', and (ii) green band trailers with ``appropriate audience'' whose associated movie is rated as G and PG.
\item \textbf{Red-band trailers:} all red-band trailers, these include restricted and mature audiences (not appropriate for children). 
\end{enumerate}

\begin{table}[]
\centering
\resizebox{0.99\columnwidth}{!}{
\begin{tabular}{|c|c|c|}
\hline
   \textbf{Green Band Trailers} & \textbf{Red Band Trailers} & \textbf{Total Trailers} \\ \hline
  1,040        &403         & 1,443                \\
\hline
\end{tabular}
}
\caption{Dataset statistics~\label{dataset}}
\end{table}

We also extracted separate audio files and trailer subtitles. Subtitles include narrator and actor speech.  Some of the YouTube trailers include the video subtitle. For these cases, we pre-process the subtitles by removing timestamps to keep only words. For trailers that do not include a subtitle file, we use a python speech recognition tool \cite{WinNT} to automatically generate the subtitle from the audio. Our dataset includes 11G of audio streams. For each trailer the audio file is a combination of background music and vocals together, so the duration of audio is the same as the duration of the trailer. The number of total words in all trailer scripts is equal to 1,478,139 (on average, there are 576 words per trailer). Note that 20,783 words of the vocabulary set are unique words.

Moreover, we provide metadata of the movies themselves, and this includes IMDB-id of the movie, title, genre, name of actors and directors, and link to the poster image. Table \ref{dataset} shows the statistics of our dataset. 

\section{Methodology}
\begin{figure*}[t]
\centering
\includegraphics[width=0.8\linewidth]{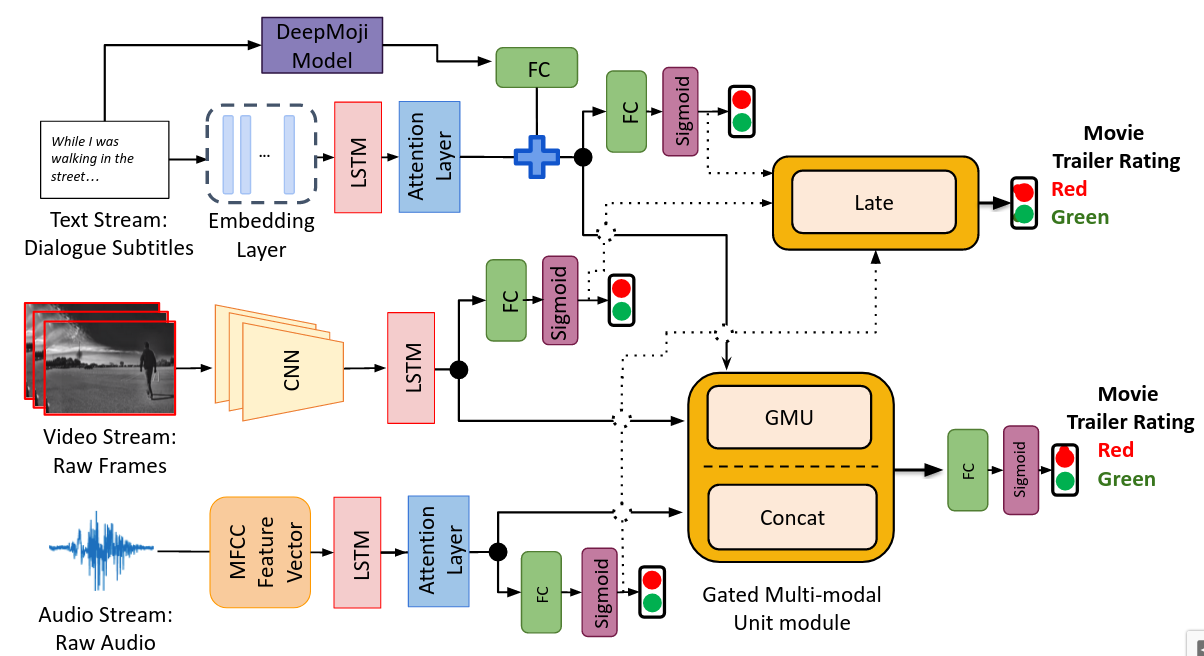}
\caption{Overview of the deep learning movie rating system and comparison of fusion methods: (i) A video subtitle is transformed into a vector representation using an Embedding Layer and then forwarded to an LSTM network with an attention layer. We concatenate the output of the attention layer with the feature vector from the DeepMoji Model. (ii) A video volume is passed through a CNN-LSTM model that is used as a feature extractor, in order to obtain a single vector representation of the entire video. (iii) Raw audio signal from the video is represented as a sequence of MFCC feature vectors, passed to an LSTM layer. (iv) Lastly, information from all modalities is combined using one of the following fusion methods, namely Gated Multi-modal Unit (GMU), Late Fusion and Feature Concatenation Fusion, before labeling the age-suitability of each trailer. FC in the diagram stands for a fully connected layer.
}
\label{overall model}
\end{figure*}

Our goal is to predict the age-suitability rating for movie trailers following the guidelines of the Advertising Administration of MPAA for trailer rating. The problem is formulated as a binary classification task where trailers are labeled as either appropriate for all audiences (green-band trailers) or restricted audiences (red-band trailers). To achieve this goal, the Multi-modal Movie Trailer Rating (MMTR) system is proposed. Within this system, the trailers are modeled as a fusion of three modalities: subtitles, audio, and video of the trailers. We train Recurrent Neural Networks (RNNs) for subtitles and audio, and a combination of Convolutional Neural Networks (CNNs) with LSTM for video, as separate streams in order to extract a representation for each respective modality. Then, we combine all stream representations using a fusion module to take advantage of the cues coming from different modalities. Figure \ref{overall model} shows the overall design for the system architecture.

Our approach is based on independently identifying the best  individual modality model and then combining information from all three monomodal models (subtitle, audio, and video) through one of the following three fusion methods: (i) Gated Multi-modal Unit (GMU) \cite{arevalo2017gated}, (ii) Feature Concatenation Fusion, or (iii)  Late Fusion. All modules of the system are described in the following sections.

\subsection{Text Stream}
The subtitles of the trailers are a rich source of information. They can help in identifying the topic of the video content. Moreover, the presence of specific words in the dialogue can be a strong indicator for some types of sensitive content, while more subtle cues can be inferred from analyzing the entire transcript. To model the information originating from the subtitles, we feed them to the following modules: 

\noindent\textbf{BERT $+$ Long Short-Term Memory (LSTM) with Attention: }
 We use BERT \cite{devlin2018bert} to leverage the well-known power of transformer-based word representations. The word vectors are then passed to an LSTM layer to model the sequence of the words in order to extract the semantic information of the text. Afterwards, the resulting hidden representation of the LSTM is passed to an attention mechanism \cite{bahdanau2014neural} to find the importance of each word in the dialogue. Even though BERT has seen a series of improvements (RoBERTa \cite{liu1907roberta}, ALBERT \cite{lan2019albert}), our goal in this paper is to present empirical evidence that a multi-modal approach can solve this task with acceptable performance, of less relevance is the specific contextualized embeddings used.

\noindent\textbf{Emotion Vector: }
We expect to observe that strong negative emotions (fear, anger, sadness) correlate more with red band trailers. Similarly, positive emotions, such as joy, are more correlated with green band trailers. We made this assumption following research by \cite{shafaei2019rating} where they found promising results for using emotions in a movie rating task.

We model emotions with the use of the DeepMoji model \cite{felbo2017using}.
%To extract the emotion in the text, the DeepMoji model \cite{felbo2017using} is employed. 
This model was trained using 1.2 billion tweets with emojis to understand how language is used to express emotions. Recent work in abusive language detection shows promising results from using DeepMoji \cite{safi2019attending}, thus it seems reasonable to expect good results in this task as well. To incorporate this model into our system, the last hidden layer representation of the pretrained model was used to transfer the text to emotional feature vectors.  Finally, the emotion vector was concatenated with the output of the attention and  the entire vector was passed to a fully connected layer to further fine-tune the joint representation. 

\subsection{Video Stream}
The video modality is a rich source of visual and temporal cues that are useful for analyzing multimedia content. Specifically, in this task, video can help for modeling the objectionable content such as a depiction of nudity or bloody scenes and suggestive elements. 
To this end, in order to learn spatiotemporal video features,  a CNN-LSTM model based on the works by \cite{lrcn2014} and \cite{Ng_2015_CVPR} is adopted. Each video is sub-sampled to a fixed number of frames, evenly distributed across its duration to form a visual temporal sequence. The raw RGB frames are used as input to a CNN model. This CNN model produces a feature representation for spatial information within each frame. The output of the final pooling layer of the CNN is passed to an LSTM that models temporal dependencies between frames. 

\subsection{Audio Stream}
The audio of the trailer can help the model to learn the genre and theme of the movie, and as a result, it is a powerful tool to distinguish red-band trailers from green-band ones. For example, \textit{horror} and \textit{thriller} movies (that usually include suspenseful music) are less likely to be suited for children. In addition to the music score, the emotion conveyed by the speakers' tone and pitch can provide relevant cues for rating the trailer. It should be noted that the entire audio is used in our model (the music and dialogue combined). To model the audio, the Mel Frequency Cepstral Coefficients (MFCC) are extracted from the audio stream. MFCCs are one of the most common feature representations for audio classification  \cite{anden2011multiscale} and speech recognition tasks \cite{tiwari2010mfcc}. The entire audio is divided to $n$ chunks, $n \in \{10,20,50,100\}$, then the MFCC feature vector is extracted for each chuck. Moreover, by performing averaging over the MFCC vector in each chunk, a fixed-length representation for the entire audio, regardless of its duration is obtained. The vector is then passed to an LSTM module to model the MFCC variations during the entire video. Lastly, by adding an attention mechanism, the model learns the importance of each audio chunk and feeds the weighted average of LSTM hidden representation to a fully connected layer that helps the model to be fine-tuned for the task.

\subsection{Fusion}
The goal of the fusion module is to learn to predict the rating of the trailer by integrating evidence from the video, audio and text modalities. We evaluate three established fusion methods in order to form a unified representation for each trailer.  

\noindent \textbf{Gated Multi-modal Unit (GMU):} The GMU allows the model to learn an intermediate representation by combining the different modalities, where the gate neurons learn to decide the contribution of each modality to the intermediate representation. A great advantage of the GMU model is its ability to adjust the activation from each modality depending on the specific instance. This method is inspired by control flow in recurrent architectures. In RNN models, the recurrent units decide how much the current and previous evidence engage in building the current state. In GMUs, the activation function for building the output using different modalities is measured, in order to form a unified intermediate representation for all modalities. 

The original GMU was successfully applied to a movie dataset of plot synopsis and movie posters  to predict genre. In the original paper, the authors implemented a bimodal system (the equation is provided in the Appendix). We follow  their formulation to extend the model to include three modalities using the straightforward approach discussed in their paper. The exact formulation is shown in Equation \ref{eq:1}; where $W_{i}$, $Y_{i}$ are learnable parameters, $x_{i}$ is the feature vector for modality $i$ and $[.,.]$ stands for concatenation.

\begin{equation}
\begin{gathered}
\label{eq:1}
    h_{1} = tanh(W_{1}.x_{1})\\
    h_{2} = tanh(W_{2}.x_2) \\
    h_{3} = tanh(W_{3}.x_3) \\
    z1= \sigma (Y_{1}.[x_{1}, x_{2}, x_{3}]) \\
    z2= \sigma (Y_{2}.[x_{1}, x_{2}, x_{3}]) \\
    z3= \sigma (Y_{3}.[x_{1}, x_{2}, x_{3}])\\
    h = z_{1}*h_{1}+ z_{2}*h_{2}+ z_{3}*h_{3}
\end{gathered}
\end{equation}
\noindent \textbf{Feature Concatenation Fusion:}
One popular fusion method is generating a joint multi-modal representation through feature concatenation \cite{baltruvsaitis2018multimodal} where the representation vectors of each modality are concatenated, and the unified representation is passed through multiple hidden layers or used directly for the prediction.  

\noindent \textbf{Late Fusion:}
Another vastly used fusion method is late fusion \cite{10.1007/978-3-319-23989-7_8}. In late fusion, different modalities are merged in the decision level using various rules (e.g., majority voting, averaging) \cite{baltruvsaitis2018multimodal}. Here, the average of all modality outputs is calculated and used as the final output. 
% Two other popular fusion methods that have shown competitive results in many fusion tasks \cite{baltruvsaitis2018multimodal} are: (i) feature concatenation fusion, where the representation vectors are just concatenated, and (ii) late fusion, the average of all modality outputs is calculated and used as the final output. 

Before performing either feature concatenation or GMU  based fusion, information from each modality is represented with a feature vector extracted from pretrained models, acting as modality streams. Then, we transform the vectors from all modalities into a single vector using the GMU or concatenation module. Finally, we pass the fused representation to a fully connected layer, creating a vector of size two (we have two classes). The sigmoid function is then applied to the two-dimensional vector to assign a label to each trailer. For late fusion, we capture the output of each single modality model before the sigmoid function (vectors of size two) and compute the average. Lastly, we pass the single representation to a sigmoid function for the classification.

\section{Experiments}
The goal of this section is to demonstrate that a multi-modal approach is an effective way to solve the task. We, therefore, compare the prediction performance of single modality models against all multimodal variations of the system. We also compare the performance of the multimodal variations of  our proposed system (MMTR system) against each other.

% And, to define the validation set, we select 10\% of the train set.
As mentioned in the dataset section, the MM-Trailer dataset is imbalanced. Thus, to obtain reliable results, 5 fold cross-validation was selected as an evaluation method. In each fold, we select 10\% of the train sent as the validation set to obtain the best model.
It should be noted that the dataset was split using the stratified approach, so as to ensure that each set has the same proportion of examples from each class. The metric used to evaluate the performance is the weighted F1 score, averaged over all 5 folds for each experiment. Implementation details of the different models are provided in the Appendix. 

\subsection{Baseline Methods}
\paragraph{Most Frequent Baseline:} The first baseline is a naive approach to show that the problem is not easy to solve. In this model, we assign the most frequent class to all the instances in the validation and test sets, and we measure the F1 score by considering the ground truth label. 
\paragraph{Text Baseline - Traditional Machine Learning:}  For the text baseline model, we provide a traditional machine learning method with hand-crafted features. We extract unigram and bigram features from subtitles and apply term frequency-inverse document frequency (TF-IDF) as the weighting scheme. Then, the feature vectors are passed to an SVM model for classification. We chose an SVM model as it performed a well on the similar task of violence detection \cite{ martinez2019violence}.

\paragraph{Text Baseline - BERT + Attention + NRC:}  A popular resource to extract the emotion in the text is the NRC emotion lexicon \cite{mohammad2011once}. This dictionary maps words to eight different emotions (anger, anticipation, joy, trust, disgust, sadness, surprise, and fear) and two sentiments (positive and negative). Using this dictionary, we compute the normalized count of words per emotion over the entire subtitle and create a vector of size 10 for each trailer. We use this vector as an alternative to DeepMoji vector in the model. 

\paragraph{Text Baseline - DeepMoji + fully connected layer:}  To show how much emotion by itself can contribute to the prediction of rating,  we only use the DeepMoji vector as the input and pass it to a fully connected layer and sigmoid classifier for the prediction. 

\paragraph{Video Baseline:}

Our video baseline is based on the deep 3-dimensional convolutional network (3D CNN) architecture proposed by \cite{3DCNN}. The 3D-CNN architecture applies 3D convolution and 3D pooling operations on video volumes instead of images. Each video is sub-sampled to an 18 evenly distributed frames that are used as input to the model. The training was performed for 50 epochs, using a 0.5 dropout rate, with a learning rate of $10^{-5}$ and a batch size of eight samples.

\paragraph{Audio Baseline:} 
CNNs have shown promising results for audio classification \cite{45611}. To this end, for each full video, the log-Mel spectrogram is extracted from the audio using the LibROSA python library \cite{mcfee2015librosa} and then used as input to a CNN architecture. For the log-Mel spectrograms 128 Mel-spaced frequency bins were used, while for the CNN model for this baseline, Inception V3 was adopted. The CNN model was trained for 100 epochs using a batch size of 64 samples and a learning rate of $10^{-5}$. An early stopping policy was used during training to avoid over-fitting.

\begin{table*}[h!]
\centering
\scalebox{0.7}{
\begin{tabular}{|c|l|l|l|}
\hline
\multicolumn{1}{|l|}{}                  & \multicolumn{1}{c|}{\textbf{Model}}                      & \textbf{Val-WF} & \textbf{Test-WF} \\ \hline
\multirow{6}{*} {\makecell{Single Modality \\ Baselines}}\ & Most Frequent Baseline                                             &60.82   &60.37    \\\cline{2-4}   
                                        & Text Baseline: Traditional Machine Learning                             &76.38	  &75.02   \\ 
                                        & Text Baseline: BERT+ Attention (T-BA)    &  84.91	&81.99 \\  
                                        & Text Baseline: BERT+ Attention+ NRC & 85.45	&81.67   \\ 
                                        & Text Baseline: DeepMoji+ fully connected layer (DeepMoji+FC) & 73.19	&68.23   \\  \cline{2-4} 
                                        & Video Baseline                                  &81.82	&75.33       \\  \cline{2-4} 
                                        & Audio Baseline                                  &82.25	&72.62        \\  \Xhline{2\arrayrulewidth}

\multirow{3}{*}{\makecell{Single Modality\\Models}}\ & Audio- MFCC (A-MFCC)                              & 77.56	&73.86      \\
                                        & Text- BERT+ Attention+ DeepMoji (T-BAD)                                &86.41	&82.67*  \\ 
                                        & Video- CNN/LSTM (V-CNN/LSTM)                          &87.06	&79.41   \\ \hline \hline

\multirow{3}{*}{Late (Fusion using two modalities)}                    & T-BAD + A-MFCC   &87.85 	&82.41    \\ 
                                        & T-BAD + V-CNN/LSTM                                  &87.31 	&84.12 \\  
                                        & A-MFCC + V-CNN/LSTM                                  &84.63	&79.68    \\  \hline
                                        
\multirow{3}{*}{Concatenation (Fusion using two modalities)}                    & T-BAD + A-MFCC   &87.15 	&82.17    \\  
                                        & T-BAD + V-CNN/LSTM                                  &89.25 	&82.80 \\   
                                        & A-MFCC + V-CNN/LSTM                                  &87.41	& 78.70  \\\hline
                                        
\multirow{3}{*}{GMU (Fusion using two modalities)}                    & T-BAD + A-MFCC   & 88.08	&83.37     \\ 
                                        & T-BAD + V-CNN/LSTM                                  & 88.00	&83.34  \\   
                                        & A-MFCC + V-CNN/LSTM                                  &85.63	&80.35    \\ \hline  \hline
                                        
\multirow{3}{*}{Fusion using all Modalities (MMTR)}        & Late (T-BAD + A-MFCC + V-CNN/LSTM)           &89.88	&85.60   \\ 
                                        & Mid (Concat) (T-BAD + A-MFCC + V-CNN/LSTM)   & 89.97	&82.75   \\  
                                        
% \multirow{1}{*}{MMTR}  
                                        & Mid (GMU) T-BAD + A-MFCC + V-CNN/LSTM                         &\textbf{91.05}	 &\textbf{86.06}*         \\ \Xhline{2\arrayrulewidth}
                                        \hline
\end{tabular}}
\caption{Evaluation of the different variants of the MMTR system and other baselines using the MM-Trailer dataset. WF stands for weighted F1 score and results are averaged over 5 folds. A `*' indicates that the difference between the two classifiers' performance is shown to be statistically significant. ~\label{results}}

\end{table*}

\section{Results}
Table \ref{results} summarizes the results of our experiments. To examine the contribution of each modality for the rating task, we report the results for all single modality models; Audio only Model (A-MFCC), Text only Model with DeepMoji (T-BAD), and Video only Model (V-CNN/LSTM ). As expected, our experimental results confirm that by leveraging all modalities we achieve a better result. As noted in Table~\ref{results} the highest weighted F1 score, 86.06\%,  is achieved by the GMU Fusion variant of the MMTR model with all modalities. This result improves the weighted F1-score of the best single modality model (T-BAD) over 3 percentage points ($P < 0.05$ based on t-test).

We also report the result for different combinations of two modalities using all fusion methods to show the effect of engaging all modalities (T-BAD + A-MFCC, T-BAD + V-CNN/LSTM , A-MFCC +V-CNN/LSTM , T-BAD + A-MFCC + V-CNN/LSTM). Based on the results, the combination of two modalities works better than every single modality, yet not as good as the combination of all modalities. 

When comparing the different fusion approaches, we can see that GMU fusion outperforms the concatenation fusion systems. We speculate that the gains from GMU come from the ability of the gated unit to dynamically adapt the contribution of each modality to the intermediate representation. Statistical significance testing using t-test, demonstrated a significant difference between GMU and feature concatenation fusion ($p-value < 0.05$). However, the test does not confirm a significant difference between late fusion and GMU. Thus, we can claim that for the trailer age-suitability  problem, late fusion can generalize as good as GMU fusion.  

The results for T-BAD and T-BA indicate that DeepMoji is a relevant feature for the rating task, and it helps the model to better discriminate red-band trailers from green-band ones. However, the result of DeepMoji+FC shows that the DeepMoji model is not sufficient to solve the task.

To obtain a better understanding of fusion results, we also provide other evaluation metrics using  the MMTR system variant with GMU Fusion (as GMU version is the winner approach based on the result table). Based on the detailed result, most of the incorrectly predicted instances are red-band trailers. The first potential explanation behind this observation is that there are fewer instances of red-band trailers in our training set compared to green-band. As a result, it is more difficult for the model to capture all patterns in this class. The second reason may be the diversity of video content in red-band trailers. Recall that this class covers any content that is not appropriate for children. It is thus reasonable to assume that this class is more heterogeneous than the green band class. We plan to explore the possibility of a fine-grained classification of objectionable content as the next steps in this work.

\section{Discussion}

To analyze the weaknesses and strengths of the MMTR system, we first investigate the incorrectly predicted cases using the most effective version of the system (GMU Fusion) on each fold of the data. By averaging results over all folds, about 35\% of incorrectly predicted cases with the MMTR system are also incorrectly predicted by each and every modality independently, fusion is thus unlikely to help in this case. We found that in about 50\% of the instances where two modalities predict the wrong rating, the MMTR system justifiably trusts the single modality that is correct. In about 93\% of the cases where only one modality is wrong, the MMTR GMU Fusion variant predicts the correct label, relying on the other two modalities.

\begin{table}[!]
\centering

\scalebox{0.8}{
\begin{tabular}{ccccc} 
\hline
\textbf{Model}              & \textbf{precision} & \textbf{recall} & \textbf{F1-score}  \\ 
\hline
\multicolumn{1}{l}{Green} & 87.4\%                  &95.0 \%     &91.0 \%                                     \\
\multicolumn{1}{l}{Red} & 83.6\%                   &65.0  \%                &72.8  \%                                     \\ \hline
\multicolumn{1}{l}{Macro avg} & 85.6\%                   &79.8  \%                &82.0  \%                                     \\
\multicolumn{1}{l}{Weighted avg} &86.6 \%                   &86.4  \%                & 86.0 \%                                     \\

\hline
\end{tabular}
}
\caption{Performance of the MMTR system using alternative metrics by performing 5 fold cross-validation evaluation method. The results are averaged over 5 folds.}
\label{table:metrics}
\end{table}

After averaging results among all folds, we notice that the MMTR GMU Fusion variant system is not able to predict about 38 out of 294 instances per fold. We watched 40 incorrectly classified trailers (selected across all folds) to analyze why the model is not able to successfully predict the label.  We introduce the following hypothesis for each of the individual modalities: 

1) \textit{Text Modality:} One main source of errors in text modality comes from the output of the speech recognition tool. First, the free version of the tool only works on short audio files. As a result, we split the whole audio to 10-second chunks. Thus, it is possible that we miss some words if the audio is cut off in the middle of the word. Also, low-quality audio impacts the word recognition rate of the automatic speech recognition system, which in turn cause the model not to recognize the specific bad words present in the video or the other way around, generate bad words by mistake (detect ``please'' as ``pussy''). However, in some cases, the trailer either has very little speech (less than 10 words) or there is really no sensitive content in the language used. Not surprisingly, the text modality cannot work properly. Finally, in some green band videos, we observed that the trailer subtitles have the words ``gun'' and ``shot'', thus they are predicted incorrectly by the text modality. It seems that the text model is biased against the occurrence of these words that are presumably strongly correlated with violent content.

2) \textit{Video Modality:} One main reason that the video modality model misses the sensitive content may relate to the video sampling rate. The inappropriate/violent scenes in these trailers disappear fast, or they may appear with a low frequency. As a result, we may miss them during sampling the frames in our model. The second potential reason is the quality of the trailers. We recognized that some of the trailers are old or are available in small files, so the frames are blurry, and even in some cases, not very clear to the human viewer. Lastly, we found out, there are some green-band trailers that still include brief sensitive content like the depiction of guns and blood, and our video modality model predicts them as red-band. These instances are mostly the R-rated movies that are sanitized for the trailer. However, the theme of the movie reflects itself in some frames. We can conclude that sometimes a single rating is not sufficient for expressing the type of the content, and as future work, we can predict a list of sensitive material in the video instead of a single label. 

3) \textit{Audio Modality:} In some cases, the music of the trailer is not compatible with the content. For example, we encountered musical movies with a high level of violence, but with smooth jazz music. Thus, it is difficult for the audio modality to distinguish between appropriate and inappropriate content. Moreover, in audio modality (similarly to the video modality), we capture samples from the continuous stream. Hence, if the intense audio (such as a scream or a gunshot) happens in a short period, our model may miss that. 

We also investigated the genre of incorrectly predicted trailers in one of the data folds. The interesting point is that, for incorrectly classified red-band trailers, 55\% are categorized as ``Thriller'' or ``Horror'' movies and 30\% as ``Comedy'' (based on IMDB metadata). We do not incorporate metadata into our model to make the model suitable for any kind of online content. This observation shows that the genre of the movie can be a potential feature for the model if we have metadata available.

% \noindent\textbf{Gold Standard Versus Automatically Generated Subtitles: } As we mentioned in Section \ref{Datase},  some of the trailers come with prepared subtitle files. For the rest, we automatically generate the subtitle from audio files. To investigate the effect of  generated text quality, we provide the statistics of the incorrectly predicted red-ban trailers for both original and synthesized subtitles in the test set of one the folds. In the red-band group, approximately half of the trailers have the original subtitles, and the other half have synthesized subtitles. For the original group, only 22\% of the samples are predicted incorrectly by text modality model, while for the synthesized group 58\% are predicted incorrectly by the same model. Based on this result, we can empirically say that the quality of the transcript impacts the quality of the text-based model, and it is likely that higher quality transcripts will result in a stronger text-based model.

\section{Conclusion}
In this paper, we examined the role of multi-modality in addressing the novel problem of trailer rating classification. We present a deep learning system named MMTR for automating the task of movie trailer age-suitability rating. MMTR fuses information from the video, audio, and text modalities. We also introduce a new data set to support research in this area. This dataset contains movie trailer videos along with their rating and metadata. Several experiments using our proposed dataset demonstrated performance gains using the multi-modal systems (GMU and late fusion) compared to other baselines. 

Beyond the practical use of a binary classification system, we are interested to move to the more challenging task of detecting the type of objectionable content and introducing model explainability and interpretability elements within the MMTR System. 

\bibliography{anthology,eacl2021}
\bibliographystyle{acl_natbib}

\end{document}